\documentclass{article}
\usepackage[utf8]{inputenc}
\usepackage{amssymb}
\usepackage[table,xcdraw]{xcolor}

\usepackage{booktabs}
\usepackage{colortbl}

\usepackage{geometry}
\usepackage{changepage}
\usepackage{arydshln}
\usepackage{multirow}
\usepackage{caption}
\usepackage{tikz}
\usepackage{float}

\usepackage{multirow}

\usepackage{graphics}

\usepackage{biblatex} 
\title{A Framework and Benchmarking Study for Counterfactual Generating Methods on Tabular Data}
\author{Raphael Mazzine, David Martens}
\date{July 2021}

\usepackage{graphicx}

\begin{document}

\maketitle

\section*{Abstract}
Counterfactual explanations are viewed as an effective way to explain machine learning predictions. This interest is reflected by a relatively young literature with already dozens of algorithms aiming to generate such explanations. These algorithms are focused on finding how features can be modified to change the output classification. However, this rather general objective can be achieved in different ways, which brings about the need for a methodology to test and benchmark these algorithms. The contributions of this work are manifold: First, a large benchmarking study of 10 algorithmic approaches on 22 tabular datasets is performed, using 9 relevant evaluation metrics. Second, the introduction of a novel, first of its kind, framework to test counterfactual generation algorithms. Third, a set of objective metrics to evaluate and compare counterfactual results. And finally, insight from the benchmarking results that indicate which approaches obtain the best performance on what type of dataset. This benchmarking study and framework can help practitioners in determining which technique and building blocks most suit their context, and can help researchers in the design and evaluation of current and future counterfactual generation algorithms. Our findings show that, overall, there's no single best algorithm to generate counterfactual explanations as the performance highly depends on properties related to the dataset, model, score and factual point specificities.

\section{Introduction}
Machine learning algorithms are becoming ever more common in our daily lives~\cite{lee2020machine}. One of the reasons for this widespread application is the high prediction accuracy those methods can achieve.
However, this gain in predictive accuracy comes at a cost of increased complexity of the prediction models and in a resulting lack of interpretability~\cite{lundberg2017unified}. The inability to explain why certain predictions are made can have a drastic impact to the adoption of automated decision making in society, as people are often reluctant to use such complex models, even if they are known to improve the predictive performance~\cite{martens2007comprehensible, kayande2009incorporating, umanath1994multiattribute, limayem2000providing, lilien2004dss, arnold2006differential, angelov2018toward}. Furthermore, these models may hide unfair biases that discriminate against sensitive groups~\cite{10.1145/3194770.3194776, dunkelau2019fairness,soares2019fair,dodge2019explaining}. These problems can be even more critical when models are constantly created and updated, as often observed in real-time applications \cite{vskrjanc2019evolving}.

To solve this deficiency, multiple strategies are proposed in a research domain commonly referred to as `eXplainable AI' (XAI)~\cite{linardatos2020explainable}, aimed at unveiling the high complexity of the models obtained through machine learning methodologies as deep neural networks~\cite{samek2017explainable,gu2020highly}, ensemble methods~\cite{hatwell2021gbt, petkovic2018improving} and support vector machines~\cite{barbella2009understanding}. They also have vast application in various fields, including finance \cite{kute2021deep,demajo2020explainable}, medicine \cite{porto2021minimum,gulum2021review} and self-driving cars \cite{soares2019explainable,lorente2021explaining}. One particular methodology of our interest is the generation of counterfactual explanations.

Counterfactual explanations have a long history, even beyond that of XAI, with roots in both psychology and philosophy~\cite{sep-causation-counterfactual, ijcai2019-876, kahneman1986norm}. According to Lipton~\cite{lipton_1990} counterfactual explanations answer the question \textit{``Why P rather than Q?''}. The outcomes P and Q are termed fact and foil, respectively, where the former is the event that occurred (in this article referred as factual) and the foil is the alternate case that did not occurred (named here as counterfactual).

Several studies indicate the advantages of counterfactual explanations~\cite{binns2018s, dodge2019explaining}, as they can make algorithmic decisions more understandable for humans~\cite{byrne2019counterfactuals}. Additionally, the recent legal requirements to the use of machine learning in decision making process such as those asked in GDRP can also be fulfilled with counterfactual explanations~\cite{wachter2017counterfactual}. Counterfactual explanations additionally have key advantages over feature importance methodologies (like LIME \cite{ribeiro2016should} or SHAP\cite{lundberg2017unified}):  although the weight of each feature may relate to why the model has a certain prediction score, it may not explain the model decision, as those methodologies do not track the influence that features have on changing the output class \cite{SEDC_IMPLEMENTATION}.

All these potential benefits caused the development of a continuously increasing number of counterfactual explanation generation methodologies over the last few years \cite{keane2021if}. In recent reviews of those methodologies \cite{verma2020counterfactual, karimi2020survey}, it was found they have a wide variation in  implementation aspects, like optimization strategy, model and data access and input data compatibility. This variation, most of times, are implemented as trade-offs for desired objectives in the counterfactual generation.

The reviews found important patterns among all methodologies, for example, most of them use  heuristic rules and/or gradient optimization techniques as search strategy, are designed to work with differentiable models (such as neural networks) and are focused on tabular data. However, for the other aspects, we find less clear patterns, specially because they try to focus on specific settings (for example, generating counterfactuals without need to access the model or data) or try to have different goals (for example, generating diverse counterfactuals). This variation and complexity in the generation of counterfactuals may be one reason why no benchmarking study with different algorithmic approaches has been conducted in the literature yet~\cite{karimi2020survey}. However, such study is of fundamental importance to explore which algorithmic implementations are preferred in specific data settings and to indicate the research directions for improved counterfactual learning.

The development of a benchmarking study relies primarily in the overall experimental setup to run different counterfactual explanation generators and the test conditions.
To compare different counterfactual explanations, clear and well defined metrics are also to be provided.  Some works~\cite{sokol2020explainability} give an overall framework to systematically analyze explainable approaches,  however, these evaluations are mostly ``high level'', without an experimental comparison, and do not provide means to effectively compare the outputs of counterfactual explanation generators.

This article aims to address this research gap and contributes in several ways in the design and implementation of benchmarks of counterfactual explanations algorithms:

\begin{enumerate}
    \item A benchmarking study of 10 counterfactual generation algorithms, on 22 datasets, assessing the algorithmic and explanation quality on 9 relevant metrics.
    \item Analysis and discussion on which algorithmic approach performs well for what dataset type. This insight can guide practitioners in choosing a suitable method.
    \item A benchmark framework implementation to test different counterfactual generation algorithms. The framework is open-source, allows to easily add new algorithms, and is intended to be a gold standard to evaluate and compare counterfactual generating algorithms.
\end{enumerate}

\section{Benchmark Components}
\subsection{Counterfactual Explanation Generation Methods} \label{counterfactualexpalgorithmsused}
Our choice of counterfactual generating algorithms is motivated by the following prerequisites: an open-source implementation, the use of novel theoretical and algorithmic strategies, and compatibility (in the code implementation) with artificial neural network models and tabular data. This resulted in 10 different algorithmic approaches to generate counterfactual explanations.
The techniques can be categorized according to (1) the search strategy, (2) required access to the training data and (3) model internals, (4) whether the method is agnostic to a prediction model or only works for a specific model, (5) whether it implements additional optimization terms, and (6) whether it allows for additional custom terms, such as enforcing one-hot encoding, as summarized in Table~\ref{tab:Frameworks}. 

Each technique follows a certain search strategy. The two most common approaches is the use of convex optimization methodologies or custom heuristic rules. Convex optimization routines will include calculation of gradients, and hence require a differentiable scoring  function. For a neural network, this requires having access to the weights~\cite{DICE, ICEGP, MLEXPLAIN, SYNAS} or using numerical estimation of gradients~\cite{ICEGP}. In case of heuristic rules, different procedures have been proposed that usually do not require to access model inner settings and, therefore, have better model compatibility. 
Note that as the method ALIBIC has two different implementations, both are included in this study. One uses the internal neural network weights to calculate gradients (we call this implementation  ALIBIC), while the other will estimate the gradients without using the internal weights, which we refer to as ALIBICNOGRAD.

Some algorithms require having access to the training data and/or the model internals. The training data requirement is generally aimed to improve counterfactual explanation quality, since the data can serve as supporting evidence in the counterfactual generation process. Such usage is implemented by ALIBIC, DiCE, GrowingSpheres, LORE, MACE, SEDC and SynAS. Similarly, algorithms like ALIBIC (although it's not mandatory), CADEX, DiCE, MACE, ML-Explain and SynAS require access to the inner model settings, such as the already cited neural network weights, to allow the use of specific optimization techniques.

Several algorithms take advantage of additional optimization terms to improve some aspects of the counterfactual explanation quality. When convex optimization techniques are used, additional loss terms can include functions that measure sparsity (the number of changes needed to generate a counterfactual), proximity (how close the counterfactual is from the factual case), feasibility (statistical similarity of the counterfactual point to the data manifold) and diversity (generate several different counterfactual explanations), these terms are discussed in more detail in Section \ref{ssec:Metrics}. 

For sparsity and improvement of the distance from the counterfactual to the factual point, ALIBIC and ML-Explain use L1 and L2 norms to calculate an elastic net loss term. And to improve both the proximity to the factual point and feasibility of changes, ALIBIC and DiCE use distance metrics, such euclidean distance, with some modifications and/or additional terms, where ALIBIC uses autoencoders trained with factual data and DiCE by using a distance metric that consider each feature variation. Finally, diversity is enhanced in DiCE by a function that considers the distance between each counterfactual explanation generated. 

Notwithstanding, these additional optimization terms are also used in algorithms that do not use convex optimization. For example, GrowingSpheres adds terms to control both sparsity (L0 norm) and proximity (L2 norm) of the solution, LORE adds a distance function (L2 norm) to control proximity and uses an genetic algorithm to improve feasibility, MACE  adds additional terms to control proximity, sparsity and diversity in the form of requirements of the satisfiability solver (SS) and uses dataset's values and ranges to enhance feasibility, while SEDC improves the feasibility of its results by calculating (and using) the mean of each feature.

Finally, some algorithms allow to include additional custom constraints to control feature changes. The constraints intend to avoid unwanted counterfactuals. One-hot encoding (OHE) is a very common example of constraint that is included in several algorithms (ALIBIC, DiCE, MACE, SynAS and LORE). This constraint requires that only one of the dummy variables corresponding to a single nominal variable can be assigned the numerical value of one. For example, a blood type categorical feature does not allow to be both O type and A type at the same time. Another common constraint to counterfactual results is the inclusion of weights that penalize changes in specified features (CA, DiCE, MACE and SynAS), and ranges (ALIBIC, DiCE, MACE and SynAS) that can constrain features to vary between specific chosen values.

\begin{table}[H]
\centering
\caption{Overall counterfactual explanation generation algorithm characteristics. Black filled circles indicate the feature is present, while white filled circles represent the feature is not present. \textbf{CO}: Convex Optimization. \textbf{HE}: Heuristic. \textbf{SS}: Satisfiability Solver. \textbf{REQ.}: Required. \textbf{NRE.}: Not Required. \textbf{OPT}: Optional. \textbf{Spar.}: Sparsity. \textbf{Prox.}: Proximity. \textbf{Feas.}: Feasibility. \textbf{Diver.}: Diversity. \textbf{W}: Weights. \textbf{R}: Range. \textbf{OHE}: One-Hot Enconding. }
\label{tab:Frameworks}
\begin{adjustwidth}{-3cm}{}
\begin{tabular}{c|ccccccccccc}
\multirow{2}{*}{\textbf{ Algorithm}} & \multicolumn{4}{c|}{\textbf{General Characteristics}}                                                                                                                                                                                            & \multicolumn{4}{c|}{\textbf{Additional Optimization Terms}} & \multicolumn{3}{c}{\textbf{Custom Constraints}}  \\
                                     & \begin{tabular}[c]{@{}c@{}}Find\\Method\end{tabular} & \begin{tabular}[c]{@{}c@{}}Data\\Access\end{tabular} & \begin{tabular}[c]{@{}c@{}}Model\\Access\end{tabular} & \multicolumn{1}{c|}{\begin{tabular}[c]{@{}c@{}}Model\\Comp.\end{tabular}} & Spar. & Prox. & Feas. & \multicolumn{1}{c|}{Diver.}         & W & R & OHE                                      \\ 
\cline{2-12}
ALIBIC~\cite{ICEGP}                   & CO                                                    & REQ.                                             & OPT.                                              & Agnostic                                                                  & \tikz\draw[black,fill=black] (0,0) circle (0.8ex);     & \tikz\draw[black,fill=black] (0,0) circle (0.8ex);     & \tikz\draw[black,fill=black] (0,0) circle (0.8ex);     & \tikz\draw[black,fill=white] (0,0) circle (0.8ex);                                   & \tikz\draw[black,fill=white] (0,0) circle (0.8ex); & \tikz\draw[black,fill=black] (0,0) circle (0.8ex); & \tikz\draw[black,fill=black] (0,0) circle (0.8ex);                                        \\
CADEX~\cite{CADEX}                   & CO                                                    & NRE.                                             & REQ.                                              & Specific                                                                  & \tikz\draw[black,fill=white] (0,0) circle (0.8ex);     & \tikz\draw[black,fill=white] (0,0) circle (0.8ex);     & \tikz\draw[black,fill=white] (0,0) circle (0.8ex);     & \tikz\draw[black,fill=white] (0,0) circle (0.8ex);                                   & \tikz\draw[black,fill=black] (0,0) circle (0.8ex); & \tikz\draw[black,fill=black] (0,0) circle (0.8ex); & \tikz\draw[black,fill=black] (0,0) circle (0.8ex);                                        \\
DiCE~\cite{DICE}                      & CO                                                    & OPT.                                             & REQ.                                              & Specific                                                                  & \tikz\draw[black,fill=white] (0,0) circle (0.8ex);     & \tikz\draw[black,fill=black] (0,0) circle (0.8ex);     & \tikz\draw[black,fill=black] (0,0) circle (0.8ex);     & \tikz\draw[black,fill=black] (0,0) circle (0.8ex);                                   & \tikz\draw[black,fill=black] (0,0) circle (0.8ex); & \tikz\draw[black,fill=black] (0,0) circle (0.8ex); & \tikz\draw[black,fill=black] (0,0) circle (0.8ex);                                        \\
Growing Spheres~\cite{GROWINGSPHERES} & HE                                                    & OPT.\footnotemark[1]                                             & NRE.                                          & Agnostic                                                                  & \tikz\draw[black,fill=black] (0,0) circle (0.8ex);     & \tikz\draw[black,fill=black] (0,0) circle (0.8ex);     & \tikz\draw[black,fill=white] (0,0) circle (0.8ex);     & \tikz\draw[black,fill=white] (0,0) circle (0.8ex);                                   & \tikz\draw[black,fill=white] (0,0) circle (0.8ex); & \tikz\draw[black,fill=white] (0,0) circle (0.8ex); & \tikz\draw[black,fill=white] (0,0) circle (0.8ex);                                        \\
LORE~\cite{LORE}                      & HE                                                    & REQ.                                             & NRE.                                          & Agnostic                                                                  & \tikz\draw[black,fill=white] (0,0) circle (0.8ex);     & \tikz\draw[black,fill=black] (0,0) circle (0.8ex);     & \tikz\draw[black,fill=black] (0,0) circle (0.8ex);     & \tikz\draw[black,fill=white] (0,0) circle (0.8ex);                                   & \tikz\draw[black,fill=white] (0,0) circle (0.8ex); & \tikz\draw[black,fill=white] (0,0) circle (0.8ex); & \tikz\draw[black,fill=black] (0,0) circle (0.8ex);                                        \\
MACE~\cite{MACE}                      & SS                                                    & REQ.                                             & REQ.                                              & Specific                                                                  & \tikz\draw[black,fill=black] (0,0) circle (0.8ex);     & \tikz\draw[black,fill=black] (0,0) circle (0.8ex);     & \tikz\draw[black,fill=black] (0,0) circle (0.8ex);     & \tikz\draw[black,fill=black] (0,0) circle (0.8ex);                                   & \tikz\draw[black,fill=black] (0,0) circle (0.8ex); & \tikz\draw[black,fill=black] (0,0) circle (0.8ex); & \tikz\draw[black,fill=black] (0,0) circle (0.8ex);                                        \\
ML-Explain~\cite{MLEXPLAIN}           & CO                                                    & NRE.                                         & REQ.                                              & Specific                                                                  & \tikz\draw[black,fill=black] (0,0) circle (0.8ex);    & \tikz\draw[black,fill=black] (0,0) circle (0.8ex);     & \tikz\draw[black,fill=white] (0,0) circle (0.8ex);     & \tikz\draw[black,fill=white] (0,0) circle (0.8ex);                                   & \tikz\draw[black,fill=white] (0,0) circle (0.8ex); & \tikz\draw[black,fill=white] (0,0) circle (0.8ex); & \tikz\draw[black,fill=white] (0,0) circle (0.8ex);                                        \\
SEDC~\cite{SEDC_IMPLEMENTATION}       & HE                                                    & REQ.\footnotemark[2]                                            & NRE.                                          & Agnostic                                                                  & \tikz\draw[black,fill=white] (0,0) circle (0.8ex);     & \tikz\draw[black,fill=white] (0,0) circle (0.8ex);     & \tikz\draw[black,fill=black] (0,0) circle (0.8ex);     & \tikz\draw[black,fill=white] (0,0) circle (0.8ex);                                   & \tikz\draw[black,fill=white] (0,0) circle (0.8ex); & \tikz\draw[black,fill=white] (0,0) circle (0.8ex); & \tikz\draw[black,fill=white] (0,0) circle (0.8ex);                                       \\
SynAS~\cite{SYNAS}                    & CO                                                    & REQ.                                             & REQ.                                              & Specific                                                                  & \tikz\draw[black,fill=white] (0,0) circle (0.8ex);     & \tikz\draw[black,fill=white] (0,0) circle (0.8ex);     & \tikz\draw[black,fill=white] (0,0) circle (0.8ex);     & \tikz\draw[black,fill=white] (0,0) circle (0.8ex);                                   & \tikz\draw[black,fill=black] (0,0) circle (0.8ex); & \tikz\draw[black,fill=black] (0,0) circle (0.8ex); & \tikz\draw[black,fill=black] (0,0) circle (0.8ex);                                        
\end{tabular}
\end{adjustwidth}
\end{table}

\footnotetext[1]{In the original paper, it theoretically has a step to generate points, hence not depending of data, however in the implementation example it uses the training and test data. In our benchmarking study we provide training data.}
\footnotetext[2]{SEDC implementation does not ask, directly, for the dataset access, however it requires the mean for each feature.}

\subsection{Datasets} \label{datasetssection}
The datasets are chosen from the public and well-known data repository (UCI Machine Learning Repository). Some dataset characteristics may lead to a preference of one counterfactual explanation generation algorithm rather than another. For example, a dataset with only categorical features may not lead to good results by an algorithm that assumes all features are continuous. Similarly, a dataset with a large number of rows may give advantage to algorithms that use the data points as evidence to generate counterfactuals. Therefore, datasets of various data types (both categorical, numerical and mixed variables), application domain, number of features, ratio of feature types, class balance and kinds of data constraints are selected (see Table~\ref{tab:Datasets}). Several of these datasets are widely used in counterfactual explanation studies like Adult~\cite{DICE, MACE, sharma2019certifai, white2019measurable, LORE, yousefzadeh2019interpreting, chapman2019emap, mahajan2019preserving}, Statlog - German Credit~\cite{DICE, LORE, SYNAS, CADEX}, Breast Cancer Wisconsin (BCW)\cite{white2019measurable, artelt2019efficient, yousefzadeh2019interpreting, artelt2020convex, ICEGP} and Wine~\cite{rathi2019generating, lucic2019actionable, artelt2020convex}. Note that the number of datasets included in the papers where these algorithms have been proposed range from 1 to only 4 \cite{DICE}, which further motivates the need for a large-scale benchmarking study.

\subsection{Models}
\label{sec:MethdologyModels}
The artificial neural network models are built using TensorFlow~\cite{tensorflow2015-whitepaper} and the Keras package~\footnotemark[3] \footnotetext[3]{Official website: https://keras.io/} in Python. The model architecture is chosen to optimize (validation) model performance, and includes one hidden layer, following the universal approximation theory~\cite{hornik1991approximation}.
A grid-search methodology is applied to find the optimal number of neurons~\cite{sheela2013review}, where the maximum number of neurons is twice the number of the input vector plus one ($2*(input\_vector\_size)+1$)~\cite{MAHDAVI2015407}. The grid also includes 4/5, 3/5, 2/5 and 1/5 of the maximum value. Other hyperparameters, as learning rate and number of epochs, are also chosen using a grid-search.  All other hyperparameters are set to the default values from TensorFlow/Keras package. The model with best AUC on the validation set was selected for each dataset.
The full model specifications  can be found in the supplementary material.

\begin{table}[htb]
\begin{adjustwidth}{-1.5cm}{}
\centering
\caption{Dataset Characteristics - Table with summarized information about each dataset used in the benchmarking study. Black filled circles indicate the feature is present, while white filled circles represent the feature is not present.}
\label{tab:Datasets}
\begin{tabular}{l|ccrrrrcc}
                & \multicolumn{2}{c|}{Dataset Information} & \multicolumn{4}{c|}{Feature Information}            & \multicolumn{2}{c}{Realistic Constraints}  \\ 
\hline
Soybean (small)~\cite{ucidatasets} & COMP.     & CAT.                         & 47       & 35   & 0    & 4                          & \tikz\draw[black,fill=black] (0,0) circle (0.8ex);      & \tikz\draw[black,fill=black] (0,0) circle (0.8ex);                          \\
Lymphography~\cite{ucidatasets, lymphographydataset}   & COMP.     & CAT.                         & 148      & 18   & 0    & 4      & \tikz\draw[black,fill=black] (0,0) circle (0.8ex);      & \tikz\draw[black,fill=black] (0,0) circle (0.8ex);                          \\
Hayes-Roth~\cite{ucidatasets}      & SOCI.     & CAT.                         & 160      & 5    & 0    & 3                          & \tikz\draw[black,fill=black] (0,0) circle (0.8ex);      & \tikz\draw[black,fill=black] (0,0) circle (0.8ex);                          \\
Balance Scale~\cite{ucidatasets}   & SOCI.     & CAT.                         & 625      & 4    & 0    & 3                          & \tikz\draw[black,fill=black] (0,0) circle (0.8ex);      & \tikz\draw[black,fill=black] (0,0) circle (0.8ex);                          \\
Tic-Tac-Toe \footnotemark[4]~\cite{ucidatasets}    & COMP.     & CAT.                         & 958      & 9    & 0    & 2             & \tikz\draw[black,fill=black] (0,0) circle (0.8ex);      & \tikz\draw[black,fill=black] (0,0) circle (0.8ex);                          \\
Car Evaulation~\cite{ucidatasets}  & N/A       & CAT.                         & 1,728     & 6    & 0    & 4                          & \tikz\draw[black,fill=black] (0,0) circle (0.8ex);      & \tikz\draw[black,fill=black] (0,0) circle (0.8ex);                          \\
Chess \footnotemark[5]~\cite{ucidatasets}           & GAME      & CAT.                         & 3,196     & 36   & 0    & 2            & \tikz\draw[black,fill=black] (0,0) circle (0.8ex);      & \tikz\draw[black,fill=black] (0,0) circle (0.8ex);                          \\
Nursery~\cite{ucidatasets}        & COMP.     & CAT.                         & 12,960    & 8    & 0    & 5                           & \tikz\draw[black,fill=black] (0,0) circle (0.8ex);      & \tikz\draw[black,fill=black] (0,0) circle (0.8ex);                          \\

Iris   ~\cite{ucidatasets}        & LIFE      & NUM.                         & 150      & 0    & 4    & 3                           & \tikz\draw[black,fill=black] (0,0) circle (0.8ex);      & \tikz\draw[black,fill=white] (0,0) circle (0.8ex);                            \\
Wine~\cite{ucidatasets}            & PHYS.     & NUM.                         & 178      & 0    & 13   & 3                          & \tikz\draw[black,fill=black] (0,0) circle (0.8ex);      & \tikz\draw[black,fill=white] (0,0) circle (0.8ex);                            \\
Ecoli~\cite{ucidatasets}          & LIFE      & NUM.                         & 336      & 0    & 7    & 8                           & \tikz\draw[black,fill=black] (0,0) circle (0.8ex);      & \tikz\draw[black,fill=white] (0,0) circle (0.8ex);                            \\
CMSC    \footnotemark[6]~\cite{ucidatasets, cmscdataset}        & PHYS.     & NUM.                         & 540      & 0    & 18 & 2  & \tikz\draw[black,fill=black] (0,0) circle (0.8ex);      & \tikz\draw[black,fill=white] (0,0) circle (0.8ex);                            \\

BCW \footnotemark[7]~\cite{ucidatasets}            & LIFE      & NUM.                         & 569      & 0    & 10   & 2             & \tikz\draw[black,fill=black] (0,0) circle (0.8ex);      & \tikz\draw[black,fill=black] (0,0) circle (0.8ex);                            \\
PBC \footnotemark[8]~\cite{ucidatasets}            & COMP.     & NUM.                         & 5,473     & 0    & 10   & 5             & \tikz\draw[black,fill=black] (0,0) circle (0.8ex);      & \tikz\draw[black,fill=black] (0,0) circle (0.8ex);                            \\
ISOLET~\cite{ucidatasets}         & COMP.     & NUM.                         & 7,797     & 0    & 617  & 26                          & \tikz\draw[black,fill=black] (0,0) circle (0.8ex);      & \tikz\draw[black,fill=white] (0,0) circle (0.8ex);                            \\
MAGIC GT \footnotemark[9]~\cite{ucidatasets}        & PHYS.     & NUM.                         & 19,020    & 0    & 10   & 2            & \tikz\draw[black,fill=black] (0,0) circle (0.8ex);      & \tikz\draw[black,fill=white] (0,0) circle (0.8ex);                            \\
SDD \footnotemark[10]~\cite{ucidatasets}            & COMP.     & NUM.                         & 58,509    & 0    & 49   & 11            & \tikz\draw[black,fill=black] (0,0) circle (0.8ex);      & \tikz\draw[black,fill=white] (0,0) circle (0.8ex);                            \\

Statlog - GC~\cite{ucidatasets}    & BUSI.     & MIX.                         & 511      & 19   & 19   & 2                          & \tikz\draw[black,fill=black] (0,0) circle (0.8ex);      & \tikz\draw[black,fill=black] (0,0) circle (0.8ex);                           \\
Student Perf.~\cite{ucidatasets, studentdataset}   & SOCI.     & MIX.                         & 649      & 27   & 3    & 2          & \tikz\draw[black,fill=black] (0,0) circle (0.8ex);      & \tikz\draw[black,fill=black] (0,0) circle (0.8ex);                          \\
Internet Adv.~\cite{ucidatasets}   & COMP.     & MIX.                         & 690      & 9    & 6    & 2                          & \tikz\draw[black,fill=black] (0,0) circle (0.8ex);      & \tikz\draw[black,fill=black] (0,0) circle (0.8ex);                          \\
Default of CCC~\cite{ucidatasets, defaultcccdataset}  & BUSI.     & MIX.                         & 30,000    & 9    & 14   & 2       & \tikz\draw[black,fill=black] (0,0) circle (0.8ex);      & \tikz\draw[black,fill=black] (0,0) circle (0.8ex);                          \\
Adult~\cite{ucidatasets}           & SOCI.     & MIX.                         & 48,842    & 9    & 5    & 2                          & \tikz\draw[black,fill=black] (0,0) circle (0.8ex);      & \tikz\draw[black,fill=black] (0,0) circle (0.8ex);    
\end{tabular}
\end{adjustwidth}
\end{table}

\subsection{Realistic Counterfactuals} \label{mathematicaldefinition}
If the counterfactual generation is left unconstrained, solutions might depict situations that will simply be deemed unacceptable to the end user. We briefly touch upon this problem by defining a subspace called realistic space.

Firstly, it's important to highlight the difference between \textbf{valid} and \textbf{realistic} counterfactuals. Let's start with what a valid counterfactual means. In this article, valid counterfactuals are those that flipped the original factual prediction category. For example, consider the situation where someone is denied a credit card, a valid counterfactual would be simply an alternate situation which changes the classification result, from not approved to approved. While an invalid counterfactual is also an alternate situation returned by the counterfactual generator (with some feature changes), but that was not able to flip the original factual prediction. One could also argue that an invalid counterfactual is simply not a counterfactual. On the other hand, realistic counterfactuals are defined by how realistic the counterfactual is in the input space, and is detailed below.

\begin{figure}[H]
\centering
\includegraphics[width=\textwidth]{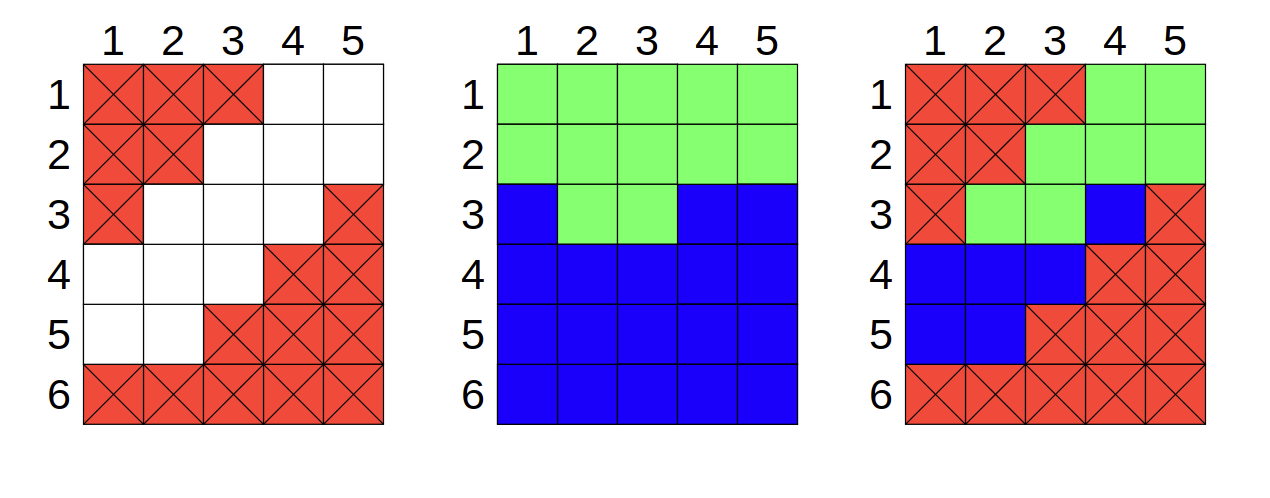}
\caption{On right: representation of the realistic space, where the white squares are the realistic states for a dataset and the red crossed squares are the unrealistic entries. On the middle: representation of a model outputs for all values, as models are just mathematical formulations, any numerical value can outputs a result (even the unrealistic dataset points). On the right: superposition of the two first, where it shows the realistic values to a classification. }
\label{fig:validspacefig}
\end{figure}

Counterfactuals correspond to datapoints (and the explanation to the difference between the factual and counterfactual). It is however possible that these datapoints are simply unrealistic, as some features may have been modified to values that are impossible in reality (Figure \ref{fig:validspacefig} illustrates this situation). One simple example is the one-hot key encoded feature, these take a categorical/nominal feature and transforms it into several binary features. This transformation implies some rules: the dummy variables must have only one ``active'' variable and they all must be binary. Re-iterating the previously mentioned example: one cannot have both type O and type A blood type, or be old and young, or married and divorced. Although the prediction model can easily provide an output for a datapoint where both dummy features are set to one (potentially even leading to a counterfactual), it is not realistic result, without a genuine representation of what this feature means to represent.

\footnotetext[4]{Tic-Tac-Toe Endgame}
\footnotetext[5]{Chess (King-Rook vs. King)}
\footnotetext[6]{Climate Model Simulation Crashes}
\footnotetext[7]{Breast Cancer Wisconsin}
\footnotetext[8]{Page Blocks Classification}
\footnotetext[9]{MAGIC Gamma Telescope}
\footnotetext[10]{Sensorless Drive Diagnosis}

The same issue occurs with other kinds of feature. In the case of numeric features, there might be an unrealistic set or range of feature values. Age for example must be larger than zero, as negative ages are not realistic entries. This shows that, although the model can accept a wide variation of numerical inputs, only a subset of them may correspond to a realistic point.

The generation of new points that follow the inner rules that define a realistic datapoint, is a challenge in itself, having own methodologies that aim to achieve it~\cite{nazabal2020handling}. This constraint has deep implications in the generation of counterfactual explanations, as if left unchecked, the counterfactual generation can lead to such unrealistic data points. It's important to remember this problem is distinct from the task of generating actionable counterfactuals~\cite{karimi2020survey}, which is much more complex~\cite{karimi2020algorithmic}. While previously we required that the counterfactual point needs to be realistic, the actionability constraint requires that the difference between the initial data instance and the counterfactual is achievable. For example, if the change implies age decreasing 5 years, the explanation is not actionable, as the change is impossible to perform, however, it can be realistic if the targeted age is larger than zero. 

The requirement to generate points that are realistic highlights the need to define a particular input space where each of features (and their combination) are conceptually genuine in relation. We call this subspace the realistic space, and will formalize this next.

Consider a set of $m$ features $\{a_1, a_2, ..., a_m\}$, the realistic space ($\textbf{R}$) is defined as a subspace  $\textbf{R} \subseteq \mathbb{R}^m$ which represents all realistic features combinations. A dataset ($\textbf{D}$) is (or should be, depending on potential data quality issues such as a negative age) a subset of the realistic space ($\textbf{D} \subseteq \textbf{R}$). If we get a point ($\textbf{x}$) in the dataset ($\textbf{x} \in \textbf{D}$) with class $y$, and a model $\textit{M}$ which transforms the point into a specific class ($\hat{y}$), we can define a counterfactual as a close-boundary point ($\textbf{c}$) where the model outputs a different class: $\textit{M}(\textbf{c})=\hat{y}$, $y \neq \hat{y}$. The realistic space can span the entire real space ($\textbf{R} = \mathbb{R}^m$) when there are no constraints and the features are free to vary. In most real world application settings however, some constraints will be imposed to keep the counterfactuals realistic. 

We therefore argue that for real-life applications, especially when involving end-users requiring explanations that make sense, a counterfactual point should belong to the realistic space ($\textbf{c} \in \textbf{R}$). This preference is additionally implied by the already mentioned fact that an unrealistic counterfactual is not supported by any data observation (as all training points are, expectedly, inside the realistic space).

To have a better understanding of how the realistic space can be defined by constraints, we can categorize these into two types of constraints: univariate and multivariate. Univariate constraints define the requirements a single feature must follow (for example, age larger than zero), while multivariate constraints describe how a set of variables, dependent of each other, must change (for example, the one hot encoding rule).

This categorization is additionally motivated by the fact that many algorithms treat these requirements differently. For example, ALIBIC, DiCE, MACE and SynAS can manually restrict the range of certain features, while ALIBIC and DiCE also use, respectively, autoencoders and statistical approaches to generate counterfactuals that are more consistent to the data manifold (thereby, hopefully, respecting both univariate and multivariate constraints). It's also worth to note that in some situations we can have both kinds of constraints, like with one-hot encoding: to ensure binary values (univariate) and single activation (multivariate).

\section{Experiments}
\subsection{Experimental design}
\begin{figure}[H]
\centering
\includegraphics[scale=0.55]{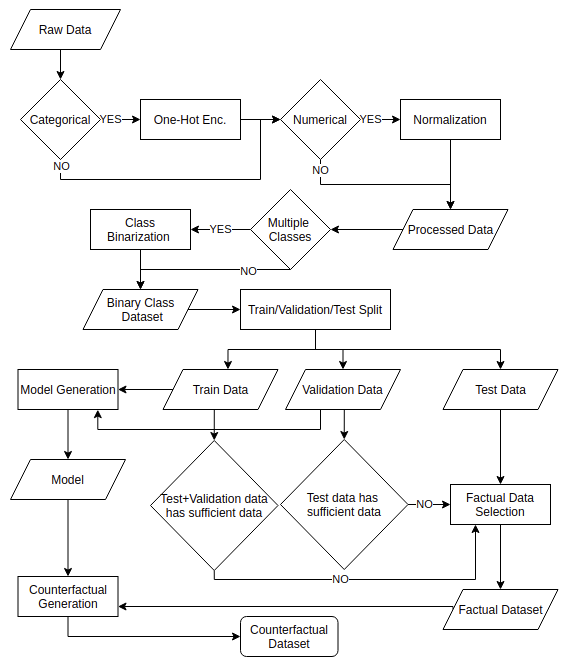}
\caption{Flowchart detailing the benchmarking process from taking the raw dataset to the counterfactual generation.}
\label{fig:counterfactualgenerationflowchartfig}
\end{figure}

The experimental design is summarized in Figure~\ref{fig:counterfactualgenerationflowchartfig}, starting with the raw data processing, normalization by using the mean ($\mu_a$) and variance ($\sigma^2_a$) of each continuous feature ($\textbf{a}_{norm}=(\textbf{a}-\mu_a)/\sigma^2_a$) and one-hot encoding of categorical features. As some counterfactual generating algorithms are not tailored towards multiclass datasets - as they, for example, don't explicitly allow to say: why class1 and not class2? But rather: why class1 and not other than class1 - the target variable is made binary for multiclass datasets, considering the majority class \textit{versus} all others. A training/validation/test split is made (60/20/20), followed by model training with definitions discussed in Section~\ref{sec:MethdologyModels}, factual data selection (the data instances to explain) and, finally, the counterfactual generation. The same models and data are used for all counterfactual explanation generation algorithms. This is an important configuration as different data processing or models can cause unreliable comparison results. Even simple modifications as scaling can lead to different distances between points which can lead to misleading results. The same can happen when working with different models, where different classification boundaries can lead to misleading results when comparing distinct counterfactual generators. To generate the counterfactual, in each dataset and class, 100 (or less, if the test set has less than 100 instances) random points are selected from the test set and used as factual points to explain. These points are the same for all tested algorithms.
All steps can be verified in the provided repository. Moreover, this benchmarking framework can accept other algorithms as it is intended to be an open benchmarking tool for counterfactual explanation algorithms.

\subsection{Metrics}
\label{ssec:Metrics}
An crucial component of this benchmarking study is defining the metrics to evaluate the explanations and the generating process.

\begin{table}
\newcolumntype{C}[1]{>{\centering\let\newline\\\arraybackslash\hspace{0pt}}m{#1}}
\begin{adjustwidth}{-1.7cm}{}
\setlength{\tabcolsep}{20pt}
\renewcommand{\arraystretch}{1.5}
\centering
\caption{Metrics used in the study with their respective descriptions.}
\label{tab:Formulasmetrics}
\begin{tabular}{ m{2em} | C{7cm} | m{6.5cm} }
Metric   & Formula & Description                                                                                         \\ 
\hline
Coverage & $H(M(\textbf{x}),M(\textbf{c})), H = \left\{ \begin{array}{ll} 0,\, if \, y=\hat{y}  \\ 1,\, if \, y \neq \hat{y} \end{array}\right.$ & Verify if the counterfactual class ($\hat{y}$) has, indeed, a different class from the factual point class ($y$) using the same model ($M$) \\
Sparsity & $\frac{1}{m}\sum\limits_{p=1}^{m}I(x_p, c_p), I = \left\{ \begin{array}{ll} 1, \, if \, x_p=c_p  \\ 0, \, if \, x_p \neq c_p \ \end{array}\right.$     & Share of features in the counterfactual array ($c_p$) that are the same as the features in the factual array ($x_p$) for all $m$ features.~                                                               \\
Stability & $I(\mathbf{c_1}, \mathbf{c_2}), I = \left\{ \begin{array}{ll} 1, \, if \, \sum\limits_{p=1}^{m}(c_{p,1}-c_{p,2}) = 0  \\ 0, \, if \, \sum\limits_{p=1}^{m}(c_{p,1}-c_{p,2}) \neq 0 \ \end{array}\right.$ & Returns 1 if the same counterfactual is returned in two different runs ($\mathbf{c_1}$ and $\mathbf{c_2}$) with same model and input data, returns 0 otherwise.\\
L2       & $\sqrt{\sum\limits_{p=1}^{m}(x_p-c_p)^2}$     & Euclidean (L2) distance between the factual point and the counterfactual point.                    \\
RUC      & $\left\{ \begin{array}{ll} 1, \, if \, c_p \in \textbf{R}  \\ 0, \, if \, c_p \notin \textbf{R} \ \end{array}\right.$     & Realistic univariate constraint, score is 1 if features are in the realistic space ($\mathbf{R}$), 0 if not.     \\
RMC      & $\left\{ \begin{array}{ll} 1, \, if \, \{c_1, c_2, ..., c_m\} \in \textbf{R}  \\ 0, \, if \, \{c_1, c_2, ..., c_m\} \notin \textbf{R} \ \end{array}\right.$     & Realistic multivariate constraint, score is 1 if set of features are in realistic space, 0 if not.  \\
MAD     & $\left\{ \begin{array}{ll} \frac{1}{m_{num}}\sum\limits_{p=1}^{m_{num}}\frac{|x_p-c_p|}{MAD_p}, \, if \, x_p \, is \,\, numerical \,\,  \\ \frac{1}{m_{cat}}\sum\limits_{p=1}^{m_{cat}}(I(x_p, c_p)), \, if \, x_p \,\, is \,\, categorical\,\, \ \end{array}\right.$     & L1 distance between the factual point and the counterfactual point divided by the feature's MAD, which is the Mean Absolute Deviation for the feature.   \\
MD       & $\sqrt{(\textbf{x}-\textbf{u})\cdot \textbf{C}^{-1} \cdot (\textbf{x} - \textbf{u})}$     & Mahalanobis Distance of the counterfactual point and the problem's dataset. Where $\textbf{u}$ is the vector with the mean values of features and $\mathbf{C^{-1}}$ is the inverse covariance matrix of the feature. \\
CT & $t_{final}-t_{init}$ & Time to generate a counterfactual in seconds.

\end{tabular}
\end{adjustwidth}
\end{table}

Coverage is a simple, but arguably one of the most important metrics, which  verifies if the provided solution actually flipped the factual class, so in other words whether a counterfactual explanation has been found.

Sparsity is a common metric used in counterfactual explanation studies as it's desirable to have short explanations, i.e. counterfactuals that differ in as few features as possible~\cite{miller2019explanation}. L2 is the euclidean distance from the factual point to the counterfactual point, it's a largely used metric for counterfactuals as, by definition, a counterfactual should be as close as possible from the factual point.

Two additional distance metrics, MAD and MD, help to understand how the algorithms generate results with respect to the dataset. The MAD distance (MAD) uses the L1 distance between the factual and counterfactual, divided by the Median Absolute Deviation (MAD). This metric is reported~\cite{wachter2017counterfactual, DICE} as being a valuable metric since it can work with different ranges across features, as it considers the variation of each changed feature. The Mahalanobis Distance (MD) ~\cite{mahalanobis1936generalized}, commonly used to find multivariate outliers, can take the correlation between features into account~\cite{kanamori2020dace, lucic2019focus}. Together, these three distance metrics measure different aspects of the solutions that might be important for specific user defined requirements. For example, if the user is only concerned about having  counterfactuals that lie as close as possible to the factual instance, with only continuous features, the Euclidean distance is an important metric to evaluate. However, if the distance is important, but it must be considered different types and ranges of features (such as those found in mixed datasets), the MAD distance can have a higher value. Finally, if the counterfactual should follow correlations between features found in data, the MD distance can be of special importance.

Stability is a score that verifies how reliable an algorithm is in generating consistent results across different runs using the same model and input data. This is calculated by comparing the results from two runs using the same parameters. A high stability means the algorithm, given the same input data and model, generates identical counterfactual results, being that a desirable characteristic for explainable approaches \cite{sokol2020explainability}. Randomisation components in the search strategy can have a detrimental effect on this metric.

Realistic scores (univariate and multivariate) are metrics which tell if the algorithm generated a counterfactual respecting the constraints defined by the dataset (consequently shows if the counterfactual is inside the realistic space). This means if equals to one, it fulfills all constraints as defined by the dataset problem. The next subsection reports on how these constraints are defined.

Finally, the last metric is the average time consumed to generate a counterfactual, measured in seconds, using the hardware and software setup described in the support resource. A full review of all metrics can be found on Table \ref{tab:Formulasmetrics}, where the formula for each individual calculation is given.

This benchmarking study does not evaluate actionability, which is a characteristic that allows a counterfactual to provide guidance on how a factual point can change the classification result by taking feasible steps. This is justified by our main focus, evaluating counterfactual explanations, which does not need to be actionable. Moreover, we consider the realistic property more critical as they are prerequisite for an actionable counterfactual explanation.
Although its usefulness providing feasible advice on how to change the factual classification, defining actionability constraints is considered a complex property~\cite{karimi2020algorithmic} because it includes time and effort considerations, while most tabular datasets provide just a static picture of feature states. Therefore, it's not possible to infer (only having the static data) how each feature can vary over time.
One solution for this problem is to design custom rules (see Table \ref{tab:Frameworks}) that constrain how each feature can vary. However, such modelling requires laborious efforts and can be error prone, which may discourage the use for real world applications.  Finally, imposing the actionability constraints itself is then again arguably a `simple' boolean requirement: does the algorithm allow to impose actionability constraints or not, for example: age can only increase (which hence relates to our realistic constraints). This further motivates why we excluded this aspect from the benchmarking study.

\subsection{Dataset Realistic Constraints}
For each dataset, one or more realistic constraints are defined to evaluate how the counterfactual explanation generation algorithms behave.

For categorical features with more than two possible values, the one-hot key encoding constraint is present, which imposes two rules: The values must be 0.0 or 1.0 (univariate constraint) and it must have just one ``activated'' entry (multivariate constraint). Categorical features that only have two values (binary features) only have the univariate constraint to check if values are realistic. For numerical features, we apply a naive univariate constraint which checks if the counterfactual feature is inside the minimum and maximum values of the feature. Therefore, values outside this range are considered unrealistic in our analysis, since those higher (or lower) values do not have supporting evidence in the dataset. The one-hot encoding features are applied for all categorical-only and mixed datasets, while the continuous feature range constraint are used for all numerical-only and mixed datasets.

Additionally, we impose three extra sets of general constraints mentioned before (all multivariate constraints) that relate to specific feature associations found in datasets. The first set is a custom constraint related to the Internet Adv. dataset, which has three features that correspond to $height$, $width$ and $ratio$ ($width/height$), therefore the constraint verifies if the relationship between features ($ratio=width/height$) is maintained. The second set of custom constraints was created for the PBC dataset, which checks the integrity of five features ($area$, $eccen$, $p\_black$, $p\_and$, $mean\_tr$) that are derived from non-linear association of five features ($height$, $length$, $area$, $blackpix$, $wb\_trans$). The last set of constraints refers to the BCW dataset, which verifies that one feature ($area$) relates to another feature ($radius$), following the equation $area=\pi(radius)^2$. A table with all information about the constraints can be found in the supplementary material.

\subsection{Algorithm comparison}\label{ssec:AlgorithmComparison}
The benchmarking framework performance ranking follows a similar statistical analysis as proposed by Demsar~\cite{demvsar2006statistical} to compare the performance of different models in multiple datasets. For each factual point and metric evaluated, we calculate the ranking of every algorithm as $r^j_i$, where $j$ is the $j$th algorithm tested and $i$ is the $i$th factual row tested. So the best algorithm will receive ranking, one, the second best ranking two, and so on. When different algorithms have the same performance, an average ranking is assigned $\bar{r}_i^{j_1,j_L}=\frac{1}{L}\sum\limits_{j=j_1}^{j_L}r^i_j$ to the $L$ algorithms that obtained the same result. The average ranking ($\bar{R}_j$) of algorithm $j$ over all  ($Q$) factuals is then obtained per dataset: $\bar{R}_j=\frac{1}{Q}\sum\limits_{i=1}^Qr^j_i$. The non-parametric Friedman test~\cite{friedman1937use, friedman1940comparison} is used to evaluate the hypothesis that all datasets perform the same. If the null hypothesis (all tests performs the same) is rejected, we proceed with a posthoc Nemenyi test~\cite{nemenyi1963distribution} to check which model(s) performed better than others.

\section{Results}
\begin{figure}[H]
\begin{adjustwidth}{-1cm}{}
\includegraphics[scale=0.48]{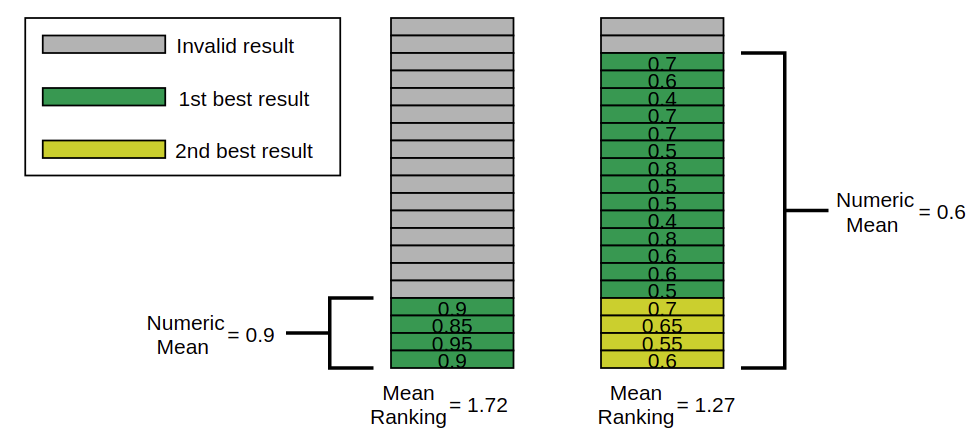}
\end{adjustwidth}
\caption{Example of how ranking and numeric results can have discrepancies. Only valid (or valid and realistic) results have values, therefore, considering that higher numerical values are better, the left set has a result numeric mean equal to 0.9, being superior to the right set (0.6). However, when calculating rankings, higher scores are considered better and being valid (or valid and realistic) is better than not, therefore, for this case we have an inverse situation, with the mean ranking of the left set being worse (1.72) than the right set (1.27).}
\label{fig:ranking_misleading}
\end{figure}

The score ranking results of each algorithm in different dataset groupings (total, numerical, categorical, mixed) are presented in Tables~\ref{tab:grouping_ranking} and \ref{tab:grouping_ranking_realistic}. The ranking analysis shows some discrepancies with numerical results, for example, an algorithm can have a better ranking in sparsity while not having the best numerical mean sparsity. The reason for this apparent paradox is the inclusion of valid (flipped the factual category) or valid and realistic (inside the realistic space) counterfactual results only in our average analysis, where invalid (or unrealistic) counterfactuals receive the lowest rankings. Therefore, an algorithm can have a better numerical result (e.g. higher mean sparsity) but as this number is obtained from few valid (or valid and realistic) counterfactuals (e.g. 10\% of total), a higher (worse) ranking is obtained, while a better ranking may have a underperforming numerical result (e.g. lower mean sparsity) but with more valid (or valid and realistic) counterfactuals (e.g. 98\% if total). This reasoning avoids  misleading results, where a numerical result may not reflect the best approach since it works for a minority of cases. Figure \ref{fig:ranking_misleading} illustrates such a situation with more details.

\begin{table}[]
\caption{Average rankings for dataset groupings. White underlined rankings highlight the best numerical result for each score and the gray cells show the algorithms that, statistically, can be considered the best. The results are rounded to one decimal place, then in some cases the best ranking can have the same numerical result as other, worse, scores. \textbf{AL}: ALIBIC, \textbf{ALN}: ALIBICNOGRAD, \textbf{CA}: CADEX, \textbf{DI}: DiCE, \textbf{GS}: Growing Spheres, \textbf{LO}: LORE, \textbf{MA}: MACE, \textbf{ML}: ML-Explain, \textbf{SE}: SEDC, \textbf{SY}: SynAS.}
\label{tab:grouping_ranking}
\includegraphics[scale=0.8]{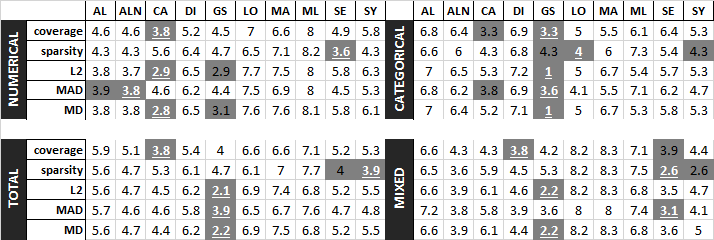}
\end{table}

In Table~\ref{tab:grouping_ranking} we evaluate all counterfactual results, regardless if they are realistic or not. Considering all datasets, we find that CADEX (CA) achieves the best coverage, finding counterfactuals for 3,549 of the 3,925 (90\%) tested factual instances, which is statistically significant better than all other algorithms. Analyzing sparsity, we have a statistical tie between two algorithms, SYNAS (SY) and SEDC (SE), where the later has a slightly better ranking with an average sparsity of 0.93 while the former has 0.87 This means that for SYNAS, on average, it must change 7\% of features while for SEDC it must change 13\% of features. For distance metrics, GrowingSpheres (GS) presents statistically better results for all rankings: for the L2 metric it has as mean distance 0.75, which is about 5.7 times shorter than the second best ranking (CA with a mean L2 of 4.3). For MAD a similar situation happens, where GS is about 2.7 times shorter than the second best ranking result, ALIBICNOGRAD (ALN). MD distance follow the two other distances trend, where GS has a mean distance value of 0.67 while CA (the second best ranking) has 4.35 (about 6.5 times larger).

We find different rankings if we consider specific data subsets: for categorical datasets for instance, we don't have CA as being  statistically best in coverage anymore, as it is able to find 1,309 counterfactuals for a total of 1,327 factual instances (about 98.6\%) while the best coverage is achieved by GS, which finds counterfactuals for 1,322 of them (about 99.6\%). However, the better numerical result for GS is not sufficient to lead to a significant difference as compared to CA, as both are statistically tied. We also have differences  in sparsity, where there's a statistical draw between three algorithms: LORE (LO), GS and SY. The numerically best in sparsity (LO) has a mean sparsity of 0.9, while GS and SY obtain values 0.59 and 0.92 respectively. The reader may notice that SY has actually the best numerical result in sparsity, a  discrepancy that was already explained earlier in this section and relates to the share of realistic counterfactuals generated. For distance metrics, GS remains the statistically best for two metrics: L2, which is the best having a L2 mean of 0.48 while the second best (LO) has 1.51 (about 3 times larger), and MD with a mean equal to 0.22, which is almost 3.5 times better than the second best (LO).

\begin{table}[]
\caption{Average rankings for dataset groupings, but only considering results that respect the realistic constraint rules. \textbf{AL}: ALIBIC, \textbf{ALN}: ALIBICNOGRAD, \textbf{CA}: CADEX, \textbf{DI}: DiCE, \textbf{GS}: Growing Spheres, \textbf{LO}: LORE, \textbf{MA}: MACE, \textbf{ML}: ML-Explain, \textbf{SE}: SEDC, \textbf{SY}: SynAS.}
\label{tab:grouping_ranking_realistic}
\includegraphics[scale=0.8]{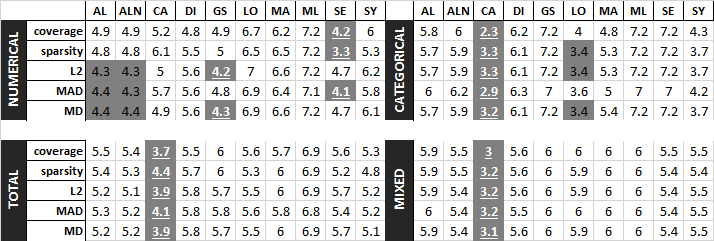}
\end{table}

For numerical datasets, CA is the statistically best in coverage, producing 1,443 counterfactuals for a total of 1,598 factual instances tested (about 90.3\%), which is significantly more than the second best algorithm (GS) which found 201 less (coverage of about 77.7\%). For sparsity, SE is the solo statistical best algorithm with an average of 0.77, however the second place (SY) has a best numerical average (0.88) but it finds 272 less counterfactual results than SE (hence explaining the lower ranking). For distance metrics, we have a less clear advantage for GS, which is among the best of two metrics, L2 and MD, but it is not the best for MAD. For L2, the two best algorithms are CA, with a mean L2 of 2.17, and GS, which average is 1.15, in this case CA has a slightly better ranking and worse numerical mean result because two reasons, first the higher number of realistic counterfactuals and second because the larger variation in results. Regarding the CA variation in L2, we see that by the larger standard deviation (equal to 2.62), while GS is much more stable (0.93). When evaluating MAD, now we see ALIBIC (AL) and ALN as being the statistically best performing algorithms, where the latter (mean MAD equal to $5.7e+5$) is slightly better than the former (mean MAD is $5.9e+5$). For MD, we have the same situation that happens to L2, with CA having a slightly better ranking and statistically tied with GS, but CA has worse mean MD (5.49) if compared to GS (1.37).

In the case of mixed datasets, CA is no longer among the best algorithms, where we verify a statistical tie between DiCE (DI) and SE. Evaluating DI, it is able to generate 896 counterfactuals from a total of 1,000 (89.6\%) while SE generates 863 (86.3\%). For sparsity, a similar situation that occurs when evaluating all datasets happens, where SE and SYNAS are considered statistically the best, but in this case SE has a slightly better ranking score. In numeric terms, SE has a mean sparsity of 0.95 which is slightly worse than SY (0.98) but SEDC's improved coverage (89.6\% versus 76.7\%) compensates this difference in ranking numbers. For distance metrics, GS performs best for L2 and MD, while SE has best results for MAD. Then, evaluating L2 we see GS has a mean distance of 0.59 which is more than 3 times lower than the second best performing algorithm, SE, which has an average L2 of 1.85. The same happens to MD, where GS has a mean value of 0.30 while SE (the second best) is about 4.8 times longer (average of 1.43). However, for MAD distance we have very different results from what we verified previously with other dataset types, where SE is the statistically best performing algorithm, being superior to the second best performing algorithm (GS) 44.5\% of times, inferior in 26.8\% cases and performs the same in 28.7\% of factual instances.

The previous results referred to a situation where counterfactuals are simply those that could flip the classification value (valid), no matter if the counterfactual is realistic or not. Then, in Table~\ref{tab:grouping_ranking_realistic} we add this additional requirement, where a counterfactual is only considered if it follow all realistic constraints the dataset has, i.e. it's valid and realistic.

Analyzing the overall results for all datasets, we already see major changes as compared to the situation where realistic counterfactuals are not required. The only similarity is that CA still performs best in terms of coverage. However, the generated number of counterfactuals dropped significantly as it only is able to produce these for 2,576 of the 3,925 factual instances (65\% of total), 27\% lower than before. The impact in coverage is even larger for the other algorithms, where for GS it generated 77\% fewer counterfactuals, SE 56\%, ALN 52\%, SY 45\%, DI 48\%, AL 39\% , MACE (MA) 23\% and LO 17\%. This has a direct impact in other scores as discrepancies between the mean numerical values and ranking scores are more salient. Evaluating sparsity, for instance, the best ranking is now obtained by CA where the mean sparsity is 0.44, but if we compare to the second best ranking result (ALN) we have a mean sparsity equal to 0.65. This situation continues for distance metrics, where for all distances CA has the best ranking, for L2, its mean distance is 4.54, while the second best (ALN) achieved 1.5, a value about 3 times better. Comparing MAD, CA has a mean distance of $4.8e+6$ and the second best (SY) has a much better score of $2.1e+4$. For MD distance, the average distance for CA is 2.89 while the second best, SY, is 0.92. Therefore, it's clear that, although CA has the best overall rankings for all metrics, this is mostly because of its superior capacity of generating counterfactuals inside the realistic space.

Considering the categorical datasets, CA remains the best or among the the best. For coverage, it has a high performance, generating 1,309 counterfactuals from a total of 1,327, this is the same value it obtained in the analysis not considering realistic constraints. For sparsity, we have a statistical tie between CA and LO, where the former has a slightly better ranking and mean sparsity of 0.6 and the latter has a mean sparsity of 0.9. In distance metrics, CA is the solo best for MAD where it has a mean of 0.40 while the second best (LO) has 0.10, which is four times better. For L2 we have a statistical draw between CA, that is slightly better in ranking and has the L2 mean equal to 3.25, and LO, which has a lower mean of 1.51. Similarly for MD, we have a statistical tie between CA and LO, where the first has a slightly lower ranking and a mean MD of 1.7 while LO has 0.78, which is also a lower value. Therefore, evaluating all metrics, except coverage, we still see the impact of CA's superior capability to generate counterfactuals in obtaining a better ranking despite inferior numerical results.

Interestingly, analyzing numerical datasets, we see a major shift in ranking results as compared to the overall results. Now CA is not among the best for any score, even for coverage. For this dataset type, coverage was the superior for SE where it produced 981 counterfactuals for a total of 1,598 factual instances (61\%). SE also is the best performing algorithm for sparsity, with a mean sparsity of 0.76, which is the third best numerical value, just behind SY (0.89) and LO (0.86). For distance metrics, there's no single best algorithm, in L2 distance for example, GS, ALN and AL perform statistically the same with mean L2 distances of 1.14, 1.27 and 1.26 respectively. For MAD distance, SE, ALN and AL are the statistically best performing in the ranking analysis where their respective mean MAD distances are $3.6e+5$, $1.6e+5$ and $2.2e+5$. AL and ALN are still among the best in MD distance together with GS, where their mean MD distance are 1.2, 1.27 and 1.14 respectively. Therefore, for distance analysis is worth to observe that AL and ALN were always among the best performing algorithms.

Finally for mixed datasets we see a trend similar to that of the overall analysis. CA performs statistically best for all scores. In coverage, for instance, it obtains the far the best result, generating counterfactual results for 597 factual cases (59.7\% of total). In comparison, the second best performing algorithm (SE) could only generate 114 counterfactuals (11.4\% of total). Similarly to what happened in previous cases, such difference in coverage will lead to discrepancies in the ranking and numerical result values. For sparsity, CA is the statistically best with a mean sparsity of 0.57 and the second best (SE) has a much better mean sparsity of 0.97. In distance metrics, CA performs as statistically best where in L2 it has a mean distance of 11.25, but the second best, SE, has a lower mean of 2.19 which is about 5 times better. For MAD CA has a mean distance of $1.9e+7$ which is much higher than the second best performing algorithm (ALN) that has $2.6e+6$ as mean MAD distance. For MD distance, a similar scenario happens where CA, the best in ranking, has a mean of 7.23 while the second best, SE, has a lower mean of 2.24.

\begin{figure}[H]
\begin{adjustwidth}{-1cm}{}
\includegraphics[scale=0.7]{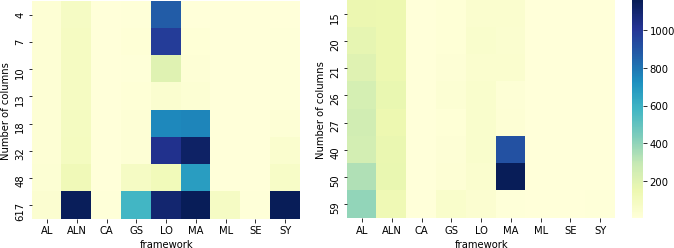}
\end{adjustwidth}
\caption{Heatmap with time (depending the number of columns) used by each algorithm to generate a counterfactual in datasets with only numerical features (left) and only categorical features (right). }
\label{fig:time_analysis}
\end{figure}

Figure~\ref{fig:time_analysis} shows how long each algorithm takes to generate a counterfactual depending the number of features in numerical and categorical datasets. We find that for numerical datasets, the algorithms ALN, GS, MA and SY tend to take more time to generate results for datasets with a higher number of features. However, LO seems to have a more unstable behaviour so it might depend on other factors as the used data generation methodology (genetic algorithms) and/or factors depending on the algorithmic implementation. Regarding the stability in time to generate counterfactuals, ML-Explain (ML) has the best standard deviation of just 0.74, closely followed by CA (1.19), AL (1.99), SY (2.74), SE (4.36), DI (4.76), GS (8.66), ALN (13.10), MA (150.69) and LO (227.89). Therefore, specially for algorithms like LO and MA, the counterfactual generation does not only depend on general dataset characteristics but also on the specific factual  point to be explained. We find, for instance, that class imbalance can be an important characteristic that defines the counterfactual generation time, as can be seen when applying the LO algorithm to the SDD dataset (where the majority class has 9 times more data than the minority): the generation of a counterfactual from majority to minority (mean 66 seconds) is 4.5 times lower than generating from minority to majority (mean 298 seconds).

In categorical datasets, AL and ALN are among the most sensitive to the number of features (see Figure~\ref{fig:time_analysis}, right chart). Although MA has higher generation times for increased number of features, it has unexpectedly low values for the dataset with a large number of categorical features (Soybean small). As reported previously, this variation can have roots in the algorithmic implementation and specific characteristics of the dataset and each factual instance. This claim is supported by the fact MA had the highest standard deviation for this kind of dataset, where ML has again the lowest value (0.49), followed by SE (0.51), SY (0.66), CA (0.66), LO (2.30), DI (3.71), GS (6.63), ALN (15.08), AL (15.18) and MA (215.55). It's worth to observe that the generation time variation for LO is much lower for categorical features (about 99 times lower), hence being another evidence that data characteristics can highly influence the counterfactual generation time in specific algorithmic implementations.

\begin{figure}[H]
\begin{adjustwidth}{-0.5cm}{}
\includegraphics[scale=0.5]{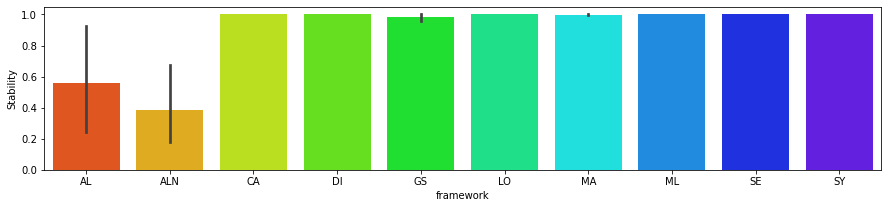}
\end{adjustwidth}
\caption{Mean stability of each algorithm for all datasets. The gray lines show the minimum and maximum value reached analyzing it for each dataset type (numerical, categorical, mixed). The algorithms CA, DI, LO, ML, SE and SY have maximum stability for all datasets. While AL has minimum stability for mixed datasets and maximum stability for numerical, ALN has minimum stability for numerical datasets and maximum stability for categorical datasets, GS has minimum stability for mixed datasets and maximum stability for numerical and categorical datasets and MA has minimum stability for numerical datasets and maximum stability for categorical and mixed datasets}
\label{fig:stability_analysis}
\end{figure}

This benchmarking study also evaluates the algorithm stability (Figure~\ref{fig:stability_analysis}), which is the generation of the same results by giving the same feature inputs over different runs. We observe that for most algorithms - namely CA, DI, LO, ML, SE and SY - maximum stability was achieved, meaning it always returns the same counterfactual for a given input. However, algorithms MA, GS, AL and ALN have problems to return the same results. For MA, it has full stability for all mixed and categorical datasets, it also has maximum stability for all numerical datasets except for BCW, where MA could not return the same result for just one factual example. Since the instability for MA is just for one factual case over all others, it still can be considered a high stability algorithm. However, the situation is different for other algorithms. GS for example returns the same result for all numerical and categorical datasets, yet provided 32 different results for mixed datasets, making the overall stability of the algorithm to be equal to 99\%. The most unstable algorithms are AL and ALN, where the former could generate the same results for only 52\% of categorical, 92\% of numerical and 24\% of mixed datasets. ALN has very low reproducible results for numerical datasets, where only 18\% could be generated in different runs, and it also gets low (but better than AL) stability for categorical (68\%) and mixed (30\%) datasets. For this analysis, it's important to highlight that AL, ALN, GS and MA do not have the option to generate reproducible results such as random seeds.

\section{Discussion}
One of the biggest challenges faced by most algorithms is to generate a counterfactual for every factual case. CA is the best overall performing algorithm in coverage, still, it could not generate counterfactuals for all factual instances. Although the algorithms' theoretical background may explain such low performance, we cannot exclude issues regarding the algorithmic implementation and fine tuning of available parameters (since we always use default settings). As discussed earlier, this has a direct impact on the ranking analysis, because not generating a counterfactual is considered the worst scenario case. Therefore, several algorithms actually have better scores (in sparsity and distance metrics) but they obtain a low ranking because of their low coverage. Having a higher coverage, then, could rearrange this ranking and, most importantly, gives more trust in the algorithm, since very low coverage like we observe for ML can refrain the user to adopt such implementations.

\begin{figure}[H]
\begin{adjustwidth}{-1.2cm}{}
\includegraphics[scale=0.4]{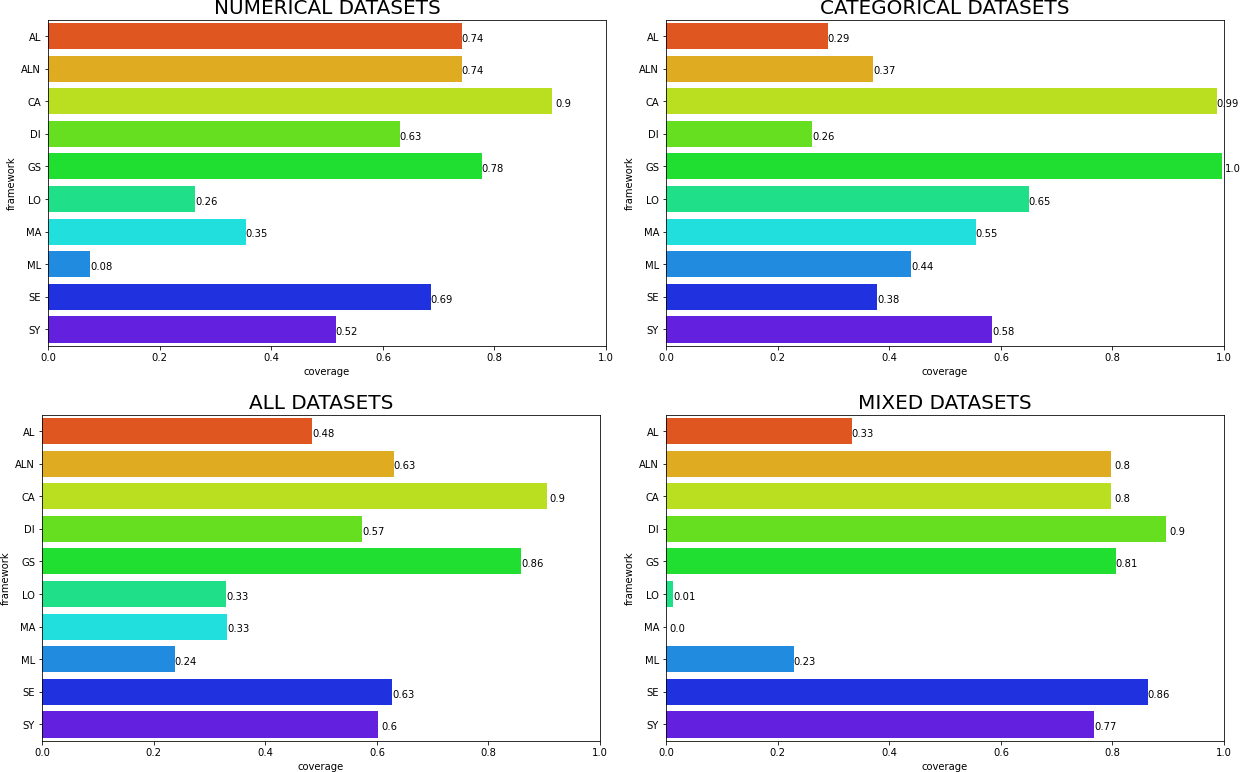}
\end{adjustwidth}
\caption{Proportion of factual examples that generates counterfactual explanations. Each chart represent a dataset type, the bottom left chart analyses all datasets used in the study.}
\label{fig:validity_analysis}
\end{figure}

We can see in Figure~\ref{fig:validity_analysis} that the coverage also depends on the  dataset type, as some algorithms can perform better for a specific kind of data but not for others. This reinforces the idea that each algorithm can have preferred data characteristics that makes it more efficient as compared to another algorithm. This becomes even more complex if we consider other scores, where the ranking analysis also shows that algorithms can have a better performance for some scores in specific data settings. For example, if we consider counterfactuals that are not realistic, GS is among the best algorithms for MAD distance in categorical datasets, but the same does not happen if we consider numerical or mixed datasets. However, this does not happen with L2 or MD distance, where GS is always among the best.

\begin{figure}[H]
\begin{adjustwidth}{-1.2cm}{}
\includegraphics[scale=0.4]{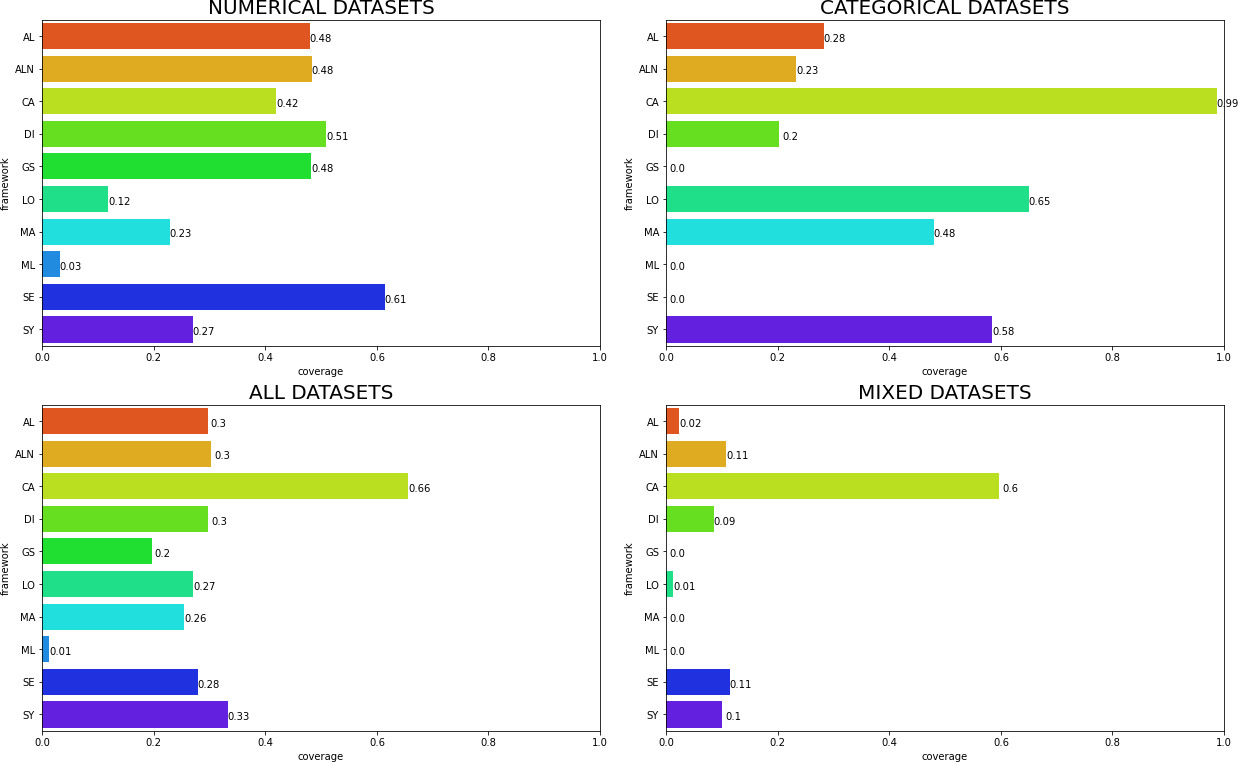}
\end{adjustwidth}
\caption{Coverage of algorithms that follows all realistic constraints. Each chart represent a data type, while the bottom left chart analyses all datasets used in the study.}
\label{fig:realistic_analysis}
\end{figure}

The situation becomes even more critical if we consider the percentage of results that followed realistic constraints. Figure~\ref{fig:realistic_analysis} shows most algorithms are far from following such constraints, where the overall best is CA, which  could generate counterfactuals for 66\% of factual instances, while no other algorithm could achieve more than 33\%. This low performance is specially worse for mixed datasets, where only CA could generate a fair number of counterfactuals (60\%) while other did not achieved more than 11\%. The direct impact of these results is that most algorithms (excluding CA for categorical datasets), can not be trusted to generate counterfactual explanations that are inside the realistic space.

When analyzing the coverage drop for numerical datasets, SE stands out as the best algorithm, which has a small reduction in performance (about 8\% less) as compared to the non-constrained case. This happens because this algorithm highly relies on dataset characteristics, where using the mean to replace feature values takes advantage of the nature of such kind of feature where a normal distribution is frequent.

It's also important to note the impact of a very common constraint found in tabular data, the one-hot encoding process, where algorithms that do not implement any theoretical and/or algorithmic approach to handle it (GS, ML , SE) perform poorly  in categorical and mixed dataset types, where such a constraint is present. This is specially noticeable for GS where it has the best ranking scores (for distance metrics) and good coverage for counterfactuals without realistic constraints, but it has a high performance drop when such constraints are applied. This can be explained because GS (and also SE and ML) treats all feature variables as numerical continuous (then, possibly breaking the univariate constraint of being binary) and they do not keep track of the set of features each one-hot encoded category has, breaking the multivariate constraint of having just one encoded feature activated (equal to 1).

Additionally, in the realistic counterfactual generation, if we analyze the results for numerical datasets, we can see a slightly better performance of algorithms that used the dataset in their generation process (SE, DI, GS, AL, ALN), although this is not a clear trend as some algorithms that also used the dataset (LO, MA and SY) performed worse than CA, which does not require the dataset in its generation process. In this case, it's understandable that counterfactual characteristics may be more compatible to inner dataset constraints when using features as supporting evidence, although there are real world application cases where providing the dataset is not an option (i.e. the dataset is private or not accessible). In those cases algorithms like CA have a clear advantage.

As previously mentioned, algorithms have different performances for different scores in distinct input data characteristics. Such high complexity means that, for specific data settings some algorithms can outperform others that are considered the overall best. For example, CA is considered to have the best overall coverage in realistic counterfactuals, but for a specific subset (numerical datasets) SE is actually the best in coverage and, additionally, although SE is the best in coverage for numerical datasets it is not for L2 and MD distances. Therefore, with the objective to improve our insights of which specific situations an algorithm perform better than the others, we use all counterfactual results to generate decision trees that highlight the best approaches depending dataset and factual instance characteristics. The decision trees settings can be found in the supplementary material.

\begin{figure}[H]
\begin{adjustwidth}{-1.55cm}{}
\includegraphics[scale=0.38]{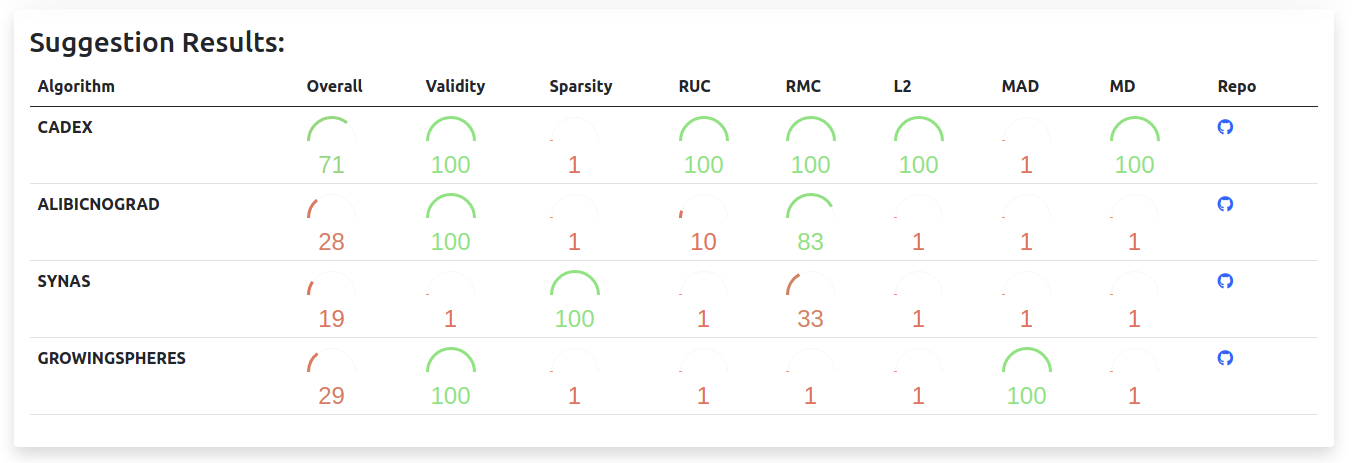}
\end{adjustwidth}
\caption{Proposed user interface to the suggestion system which tells the best performing algorithms for a specific dataset setting. The scores are calculated by using decision trees and reflect how well (compared to the others) the algorithm performed.}
\label{fig:api_example}
\end{figure}

In this analysis we find some interesting patterns, confirming that selecting the best algorithmic approach can depend on several parameters that includes characteristics from the model (like number of neurons, AUC score), dataset (number of categorical or numerical columns) and/or the factual point (prediction score, balance share of factual point class).

Considering the counterfactual explanations that do not need to be realistic, we find for categorical datasets that although GS, LO and SY are the best ranked, if we consider a large number of categorical columns (more or equal to 45) and models that have a large number of neurons (bigger than 10.5) SE is actually the best algorithm by far (in sparsity) if the factual point class is highly unbalanced (share of the factual class larger than 0.8). It's important to notice that this observation involved characteristics from the dataset (number of categorical columns), model (number of neurons) and factual point (factual point class share), this highlights that, although there are overall best algorithms, specific settings can achieve better performance with other implementations.

Bearing that in mind, it's proposed that such decision trees can be used to suggest end-users which algorithm they should use for their specific problem. Such implementation can output results like the one of Figure \ref{fig:api_example}, where a certain performance value for each algorithm in every metric is provided. The user can then select the one that best suits their requirements. Notwithstanding, the decision tree suggestions do not reduce the importance of the overall ranking, which may be better suited both to end-users that are not sure of which particular setting they will work and to researches that want to compare the performance of their algorithms.

\begin{figure}[H]
\begin{center}
\includegraphics[scale=0.4]{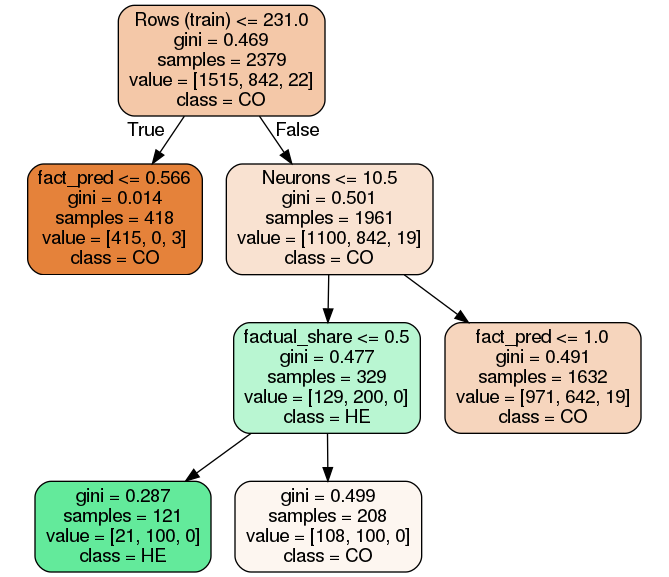}
\end{center}
\caption{Decision tree to decide which algorithm implementation pattern performs best in realistic counterfactual generation using categorical data. In this case, this decision tree evaluates the best performing in L2 distance. On top of each node, the splitting condition is shown, which includes the following information: Rows (train) - number of rows in the training set, fact\_pred - factual instance prediction score, Neurons - the number of neurons of the NN model, factual\_share - the datset percentage of the factual point class. In second on the node, there's the Gini coefficient and, below it, the number of total samples analyzed. As penultimate, we see the values for each of the three classes: first it's CO, which represents convex optimization based methods (AL, ALN, CA, DI, ML, SY), in second there's HE that represents the heuristic based methods (GS, LO, SE) and finally the SS based method (MA). Last in the node, there's an indication of which class is the majority, hence which type performs best.}
\label{fig:dt_realistic_cat_l2}
\end{figure}

This kind of analysis using decision trees can be expanded to evaluate how  implementation choices, such as search strategy, model access and data usage affect scores in different types of dataset. For example, evaluating realistic counterfactuals and categorical datasets we obtain the decision tree presented in Figure \ref{fig:dt_realistic_cat_l2}, where the first split node is related to the number of instances the dataset has and it totally separates heuristic rules from convex optimization methods (and in a minor extent SS). This can be related to how the heuristic rules work, where the dataset usage is not just to enhance loss term parameters like in convex optimization, but they actively rely on evidence from the dataset to generate counterfactuals. Therefore, not having many instances can lead to substandard performance. Although Figure \ref{fig:dt_realistic_cat_l2} just shows the decision tree for L2 score, the same behaviour is observed for MAD, MD, sparsity and coverage scores.

Another important observation obtained from the higher performance achieved by CA, specially on realistic counterfactuals, is the fact it does not have any additional loss term applied (as we only used the default settings) and no access to the dataset. We envision there is a room for improvements in generating counterfactuals for all kinds of data and those that follow realistic constraints, and the already mentioned theoretical and/or algorithmic implementation may be further improved to achieve such performance. We also highlight the characteristics of being a realistic counterfactual is highly desirable since results absent of this property only provide a mathematical justification rather than a logical reasoning for the factual instance classification. Therefore, as most algorithms underperform on this condition, this should be a goal to pursuit.

Two additional properties were evaluated by this study, stability and time to generate the counterfactual, although they are not directly associated with the counterfactual quality they may be important for real-world applications where such properties can have an important weight. For stability, in some circumstances, it might be a required characteristic~\cite{sokol2020explainability} as the user may not want to receive different results in different runs (for the same input). Although it can be argued that such instability brings more diversity to explanations, this should not be done in an uncontrolled way, hence the use of strategies like providing a random seed setting could be used to circumvent this. This is notably critical for AL and ALN, which presented very low stability as compared to other algorithms.

In terms of time requirements, some users may have limited time to give an answer (real time applications) or it can also have resource limitations, where large processing times may lead to an incompatibility with large scale applications. Additionally, it must be noted we didn't use very large datasets, which are becoming common in real-world scenario. Therefore the time to generate a counterfactual must be a common concern and those algorithms that had poor performance (like AL, ALN, MA, LO, GS and SY) may face challenges in adoption on those specific cases.

Taking all the previous observations into account, we conclude, today, there's no single `silver bullet' for counterfactual generation. End-users may evaluate their specific requirements based on metrics like the ones we presented in this article to select which algorithmic implementation it will mostly correspond to their expectation. And researchers may use this benchmarking study to compare novel approaches and test modifications of existing ones. Aiming to unveil the intrinsic complexity of several parameters and scores, the use of decision trees to suggest algorithms can also be considered a solution to the current heterogeneity in performance.

Finally, it's worth again to remember these results represent the theories and (most importantly) algorithmic implementations in a specific time, then, both theory and algorithms can be updated leading to different results that we presented here. However, our intention is also to give a starting point for the counterfactual explanation generators benchmarking, proposing a process pipeline, required theoretical foundations, scoring metrics and possible questions that will lead to the refinement of the area and possible broader adoption.

\section{Conclusions}
Benchmarking studies have fundamental value in scientific research, for example, in the case of image classification, the ImageNet Challenge~\cite{ILSVRC15} allowed us to test and compare novel methodologies that lead to advancements in the area. In the case of Counterfactual Explanations, a very recent field in XAI, our benchmarking framework aims to be beneficial in several forms. First, it allows those who are building new theoretical and/or algorithmic approaches to test their implementations more quickly and correct possible bugs or flaws in theory. Second, it provides a common ground to make an extensive analysis towards other implementations, removing any possible difference on data treatment, model design, scoring metric that might benefit one implementation over another. Third, the benchmarking study can reveal which algorithms performs best in specific conditions and, ultimately, can lead to find which specific implementation strategies work for each case.
Furthermore, as a side effect of how the benchmarking framework was built, the standardization of the counterfactual algorithm implementation, in terms of required variables and operation steps, can also benefit both researchers and end-users since this often has wide a variation among the algorithms making more difficult their understanding and application. Similarly, the use of objective scores presented in this work can also create a pattern both in nomenclature and evaluation, that often are diverse in this area.  In a future perspective, besides the obvious addition of new algorithmic implementations, the refinement of constraints definition made by domain experts, will have a positive impact on the results of this benchmarking study as it will enhance the value of realistic constraints (RUC, RMC). In the same way, despite the importance of artificial neural networks, focus of this benchmark, the addition of new model types (in particular, black box models) will give a broader compatibility to algorithms and reveal new insights on them. 

\section{ACKNOWLEDGMENTS}

Authors sincerely thanks AIFlanders for funding this research and Google for providing resources to run the benchmarks on their cloud platform.

\printbibliography
\clearpage

\begin{center}
\textbf{\Large Supplementary Material}
\end{center}

\setcounter{section}{0}
\renewcommand{\thesection}{S-\Roman{section}}

\section{Model specification}

A general overview of models is presented at Section~\ref{sec:MethdologyModels}. Below it's presented the full details for each dataset, parameters that are not explicitly indicated must be considered the default for the package. For all models the following configuration is always the same:

\begin{itemize}
    \item Number of hidden layers: 1
    \item Number of output classes: 2
    \item Activation function (hidden layer): ReLU
    \item Activation function (output layer): Softmax
    \item Optimizer: RMSprop
    \item Loss function: Categorical Crossentropy
\end{itemize}
\pagebreak

\section{Hardware used for experiments}
For the experiments, we were granted resources by Google to use their cloud computing environment. We used (for all datasets, except InternetAdv) N1 machine type, with 52 cores in a CPU platform based on Intel Skylake or later, with 195GB of memory, 400GB of SSD storage and OS Ubuntu 18.04. For InternetAdv it was used N2D machine type, using 48 cores of AMD Rome or later CPUs, with 384GB of memory and 400GB of SSD storage and OS Ubuntu 18.04.

\begin{table}[]
\begin{adjustwidth}{-1.5cm}{}
\caption{Overall data for each dataset and model. \textbf{fac0} and \textbf{fac1} are the factual datasets with original class 0 and 1 respectively. The \textbf{\% Maj.} is the share of rows of the majority class. \textbf{Col} is the number of columns, \textbf{Neu} is the selected model number of neurons in the single hidden layer, \textbf{Epc} is the number of epochs used in training and \textbf{LR} is the learning rate.}
\resizebox{18cm}{!}{%
\begin{tabular}{l|rll|cllll|lll|lllll|}
                 & \multicolumn{3}{c|}{\textbf{Rows}}                                 & \textbf{}        & \textbf{}    & \textbf{}    & \textbf{}   & \textbf{}   & \multicolumn{3}{c|}{\textbf{AUC}}               & \multicolumn{5}{c|}{\textbf{Accuracy}}                                          \\
\textbf{Dataset} & \multicolumn{1}{l}{\textbf{Total}} & \textbf{Fac0} & \textbf{Fac1} & \textbf{\% Maj.} & \textbf{Col} & \textbf{Neu} & \textbf{Ep} & \textbf{LR} & \textbf{Train} & \textbf{Valid} & \textbf{Test} & \textbf{Train} & \textbf{Valid} & \textbf{Test} & \textbf{Fac0} & \textbf{Fac1} \\
Adult            & 32,561                             & 100           & 100           & 0.76             & 107          & 215          & 50          & 0.0001      & 0.92           & 0.91           & 0.91          & 0.87           & 0.86           & 0.85          & 0.72          & 0.87          \\
BCW              & 198                                & 41            & 100           & 0.76             & 32           & 65           & 100         & 0.001       & 1              & 0.85           & 0.77          & 0.99           & 0.88           & 0.74          & 0.88          & 0.9           \\
BalanceScale     & 625                                & 100           & 100           & 0.54             & 20           & 16           & 50          & 0.01        & 1              & 1              & 1             & 1              & 0.99           & 0.98          & 0.98          & 0.99          \\
CMSC             & 540                                & 37            & 100           & 0.91             & 18           & 22           & 100         & 0.001       & 1              & 0.98           & 0.79          & 1              & 0.95           & 0.94          & 0.97          & 0.95          \\
CarEvaluation    & 1,728                              & 100           & 100           & 0.70             & 21           & 17           & 50          & 0.01        & 1              & 1              & 1             & 1              & 1              & 0.99          & 0.98          & 1             \\
Chess            & 28,056                             & 100           & 100           & 0.90             & 40           & 64           & 500         & 0.01        & 1              & 1              & 1             & 1              & 1              & 1             & 1             & 1             \\
DefaultOfCCC     & 30,000                             & 100           & 100           & 0.78             & 90           & 72           & 50          & 0.0001      & 0.79           & 0.77           & 0.79          & 0.83           & 0.82           & 0.83          & 0.68          & 0.84          \\
Ecoli            & 336                                & 100           & 100           & 0.57             & 7            & 15           & 50          & 0.001       & 0.99           & 1              & 1             & 0.98           & 0.94           & 0.97          & 0.95          & 0.96          \\
HayesRoth        & 132                                & 96            & 36            & 0.61             & 15           & 31           & 500         & 0.0001      & 0.95           & 0.96           & 0.94          & 0.87           & 0.86           & 0.88          & 0.83          & 0.97          \\
ISOLET           & 7,797                              & 100           & 100           & 0.96             & 617          & 247          & 500         & 0.001       & 1              & 1              & 1             & 1              & 1              & 1             & 1             & 1             \\
InternetAdv      & 2,359                              & 100           & 100           & 0.84             & 1558         & 1246         & 50          & 0.0001      & 1              & 0.98           & 0.99          & 0.99           & 0.97           & 0.97          & 0.96          & 0.96          \\
Iris             & 150                                & 97            & 53            & 0.67             & 4            & 3            & 50          & 0.01        & 1              & 1              & 1             & 1              & 0.93           & 0.97          & 1             & 0.94          \\
Lymphography     & 148                                & 69            & 79            & 0.55             & 50           & 40           & 100         & 0.001       & 1              & 0.92           & 0.94          & 1              & 0.81           & 0.86          & 0.91          & 0.95          \\
MagicGT          & 19,020                             & 100           & 100           & 0.65             & 10           & 12           & 500         & 0.001       & 0.94           & 0.92           & 0.93          & 0.88           & 0.86           & 0.87          & 0.88          & 0.89          \\
Nursery          & 12,960                             & 100           & 100           & 0.67             & 26           & 10           & 50          & 0.01        & 1              & 1              & 1             & 1              & 1              & 1             & 1             & 1             \\
PBC              & 5,473                              & 100           & 100           & 0.90             & 10           & 12           & 100         & 0.01        & 1              & 0.99           & 0.99          & 0.98           & 0.97           & 0.98          & 0.91          & 0.99          \\
SDD              & 58,509                             & 100           & 100           & 0.91             & 48           & 19           & 500         & 0.001       & 1              & 1              & 1             & 1              & 1              & 1             & 1             & 1             \\
SoybeanSmall     & 47                                 & 30            & 17            & 0.64             & 59           & 23           & 50          & 0.01        & 1              & 1              & 1             & 1              & 1              & 1             & 1             & 1             \\
StatlogGC        & 1,000                              & 100           & 100           & 0.70             & 59           & 119          & 100         & 0.0001      & 0.91           & 0.83           & 0.79          & 0.85           & 0.75           & 0.76          & 0.63          & 0.85          \\
StudentPerf      & 649                                & 100           & 100           & 0.54             & 43           & 17           & 100         & 0.0001      & 0.82           & 0.88           & 0.71          & 0.75           & 0.76           & 0.66          & 0.74          & 0.66          \\
TicTacToe        & 958                                & 100           & 100           & 0.65             & 27           & 11           & 500         & 0.01        & 1              & 1              & 1             & 1              & 0.98           & 0.99          & 0.98          & 1             \\
Wine             & 178                                & 100           & 70            & 0.60             & 13           & 5            & 50          & 0.01        & 1              & 1              & 1             & 1              & 0.97           & 1             & 0.99          & 1            
\end{tabular}%
}
\end{adjustwidth}
\end{table}

\section{Decision Tree Analysis}

The decisions trees were produced using the Python package Scikit-learn \cite{scikit-learn} using a fixed random state equal to 0 and maximum depth of 3, all other parameter were set as default. The used data has 8 features that are related to model, dataset and factual point characteristics:
\begin{itemize}
    \item Model features:
        \subitem \textbf{1) neurons} - Number of neurons the model has.
        \subitem \textbf{2) auc\_test} - AUC score of the model in the test set
    
    \item Dataset features:
        \subitem \textbf{3) rows\_train} - Number of rows in the train dataset.
        \subitem \textbf{4) column\_numerical} - Number of numerical columns in the dataset
        \subitem \textbf{5) columns\_categorical} - Number of categorical columns in the dataset
    
    \item Factual Point features:
        \subitem \textbf{6) misclassified} - Tells if the original factual class was misclassified (1 if yes, 0 if not) by the model.
        \subitem \textbf{7) factual\_prediction} - Prediction probability for the factual class, ranging from (0.5 to 1.0).
        \subitem \textbf{8) factual\_share} - Percentage of rows of the factual class in the dataset.
\end{itemize}

As target feature, we selected the best performing algorithm for the specific factual data point, in case where more than one algorithm are tied as best, we created additional rows having the same input features but with different targets representing the best algorithms.

For this analysis, we do not focus in getting high accurate decision trees, we focus on the Gini coefficient which low values (close to 0) show a preference for a specific algorithm, while numbers close to 1 show there's no preferable algorithm for the split conditions.

As there are several (more than 200) decision trees generated by this benchmark, you can find all them in the official CF benchmark repository link.

\begin{table}[H]
\centering
\caption{Description of constraint types for each dataset. For univariate constraints, Feature Range sets a continuous range where a variable can be changed, Feature Binary allows some features assume only 0.0 or 1.0. For multivariate constraints, Second Order describe one or more features have a second order degree relationship between then, OHE describe the activation rule of one-hot encoding.}
\begin{tabular}{l|ll}
Dataset         & Univariate Constraint         & Multivariate Constraint  \\ 
\hline
Balance Scale   & Feature Binary                & OHE                      \\
Car Evaluation  & Feature Binary                & OHE                      \\
Hayes-Roth      & Feature Binary                & OHE                      \\
Chess           & Feature Binary                & OHE                      \\
Lymphography    & Feature Binary                & OHE                      \\
Nursery         & Feature Binary                & OHE                      \\
Soybean (small) & Feature Binary                & OHE                      \\
Tic-Tac-Toe     & Feature Binary                & OHE                      \\
BCW             & Feature Range                 & Second Order             \\
Ecoli           & Feature Range                 &                          \\
Iris            & Feature Range                 &                          \\
ISOLET          & Feature Range                 &                          \\
SDD             & Feature Range                 &                          \\
PBC             & Feature Range                 & Second Order             \\
CMSC            & Feature Range                 &                          \\
MAGIC GT        & Feature Range                 &                          \\
Wine            & Feature Range                 &                          \\
Default of CCC  & Feature Range,~Feature Binary & OHE                      \\
Student Perf.   & Feature Range,~Feature Binary & OHE                      \\
Adult           & Feature Range,~Feature Binary & OHE                      \\
Internet Adv.   & Feature Range,~Feature Binary & OHE,~Second Order        \\
Statlog – GC    & Feature Range,~Feature Binary & OHE                     
\end{tabular}
\end{table}

\end{document}